\def\year{2024}\relax
\documentclass[letterpaper]{article} 
\usepackage{aaai24}  
\usepackage{times}  
\usepackage{helvet}  
\usepackage{courier}  
\usepackage[hyphens]{url}  
\usepackage{graphicx} 
\usepackage{natbib}  
\usepackage{caption} 
\DeclareCaptionStyle{ruled}{labelfont=normalfont,labelsep=colon,strut=off} 
\usepackage{algorithm}
\usepackage{algorithmic}
\usepackage{amsmath}

\newcommand{\red}[1]{{\color{red}#1}}

\usepackage{color}
\usepackage{amsfonts}
\usepackage{comment}

\usepackage{newfloat}
\usepackage{listings}
\floatstyle{ruled}
\newfloat{listing}{tb}{lst}{}
\floatname{listing}{Listing}
\title{Towards an Adaptable and Generalizable Optimization Engine in Decision and Control: A Meta Reinforcement Learning Approach}
\author{
    Sungwook Yang\textsuperscript{\rm 1},
    Chaoying Pei\textsuperscript{\rm 2}, 
    Ran Dai\textsuperscript{\rm 2}, 
    Chuangchuang Sun\textsuperscript{\rm 1} 
}
\affiliations{
    \textsuperscript{\rm 1}Mississippi State University, Mississippi State, MS 39762\\
    \textsuperscript{\rm 2}Purdue University, West Lafayette, IN 47906\\

    sy449@msstate.edu, 
    peic@purdue.edu,
    randai@purdue.edu,
    csun@ae.msstate.edu
}

\usepackage{bibentry}

\begin{document}

\maketitle

\begin{abstract}
Sampling-based model predictive control (MPC) has found significant success in optimal control problems with non-smooth system dynamics and cost function. Many machine learning-based works proposed to improve MPC by a) learning or fine-tuning the dynamics/ cost function, or b) learning to optimize for the update of the MPC controllers. For the latter, imitation learning-based optimizers are trained to update the MPC controller by mimicking the expert demonstrations, which, however, are expensive or even unavailable. More significantly, many sequential decision-making problems are in non-stationary environments, requiring that an optimizer should be \textit{adaptable} and \textit{generalizable} to update the MPC controller for solving different tasks. To address those issues, we propose to learn an optimizer based on meta-reinforcement learning (RL) to update the controllers. This optimizer does not need expert demonstration and can enable fast adaptation (e.g., few-shots) when it is deployed in unseen control tasks. Experimental results validate the effectiveness of the learned optimizer regarding fast adaptation.

\end{abstract}

\section{Introduction}

Model predictive control (MPC) is a powerful tool for sequential decision-making problems, achieving success in many areas such as cyber-physical systems. A major challenge for sampling-based MPC variants, e.g., model predictive path integral (MPPI), is the updating rule of the controllers. The design of such updated rules will significantly decide the performance of MPC and heavily depend on manual tailoring. Unifying the MPC variants under the umbrella of the first-order optimization algorithm, dynamic mirror descent~\cite{hall2013dynamical}, learning to optimize techniques ~\cite{andrychowicz2016learning, chen2022learning} are a natural option to learn performant optimizer parameterize by general function approximations (e.g., deep neural networks) as the updating rule in MPC. Imitation learning, a popular paradigm for learning from demonstration, is applied to learn an optimizer of MPPI~\cite{sacks2022learning} with the restrictive requirement of expert demonstration. Reinforcement learning (RL)-based approaches, alleviating such requirements, are also developed for learning optimizers~\cite{li2016learning}. Moreover, to enhance the stability of learning to optimize, Lyapunov-like conditions are also incorporated~\cite{sun2021fisar}.

However, in real-world deployment, tasks can be non-stationary due to the pervasive uncertainty, perturbation, noise, and adversaries in the environment. Then the requirement of adaptation and generalization of learned optimizer for MPC arises. In other words, an optimizer should still be able to train performant controllers in unseen tasks. To this end, we propose a meta-RL~\cite{finn2017model} based approach to learn optimizers that can adapt in a few shots in testing. In his approach, a meta optimizer together with multiple local task-specific optimizers is learned simultaneously, with the former as the initializer of the latter in each round of updating. Specifically, during training under a distribution of different tasks, the local optimizer and meta optimizer obtain gradient information from its associated local task and the overall tasks, respectively. This provides the meta optimizer with the ability to quickly adapt to new (in-distribution) tasks in testing with only a few samples.



\section{Problem Formulation: Learning to Optimize for Model Predictive Control}
We consider an optimal control problem in a discrete stochastic dynamical system, $\mathbf{x}_{t+1} \sim f \left( \mathbf{x}_{t}, \mathbf{u}_{t} \right)$, with the dynamics $f: \mathbb{R}^{m+n}\to\mathbb{R}^n$, states $\mathbf{x} \in \mathbf{R}^{n}$ and the control inputs $\mathbf{u} \in \mathbf{R}^{m}$. The sequence of the control inputs $\mathbf{u}$ of the MPPI algorithm is generated by using the controller parameterized by a Gaussian distribution $\mathcal{N} \left( \mu,\; \sigma^{2} \right)$. 
To find the optimal controller, learning to optimize by RL is applied.

The parameters of the controller, $\mathbf{m}_{t} = \left[ \mu_t^\top, \;\sigma_t^\top \right]^{\top}$, are optimized by using a reinforcement learning policy $\pi_\theta(\bullet)$ parameterized by the deep neural networks $\theta$. Sampling with the current MPPI controller $\mathbf{m}_{k}$, a trajectory can be rolled out as $\mathcal{D}_k = \{\mathbf{u}_{k,t},c_{k,t}\}_{t=0}^T$ under controller $\mathbf{m}_{k}$ for $T$ steps. The rule to get the update $\Delta \mathbf{m}_{k}$ is as follows
\begin{equation} \label{eq:RL-optimizers}
	\mathbf{m}_{k+1} = \mathbf{m}_{k} + \Delta \mathbf{m}_{k}, \text{with}\ \Delta \mathbf{m}_{k} = \pi_\theta (\mathbf{m}_{k}, \mathcal{D})
\end{equation}
To update the policy parameters, we use the classic RL algorithm, e.g., policy gradient, to maximize the return as $J(\theta) = \sum_{h=0}^H \gamma^h R_h$, where $k$ is the time step in an RL episode with a horizon $H$, $R_h$ is the reward, and $\gamma\in (0,1]$ is a discount factor. The gradient of the return is 
\begin{equation} \label{eq:pm_grad_fnc}
	\nabla J(\theta) =  \sum_{k = 0}^{H} \nabla_\theta \log \pi_\theta \left(\Delta\mathbf{m}_{k}|\mathbf{m}_{k}, \mathcal{D}_k \right) \sum_{t=0, k=0}^{t=T,k=H} \frac{-c_{k,t} }{T}
\end{equation}
where the RL reward to update the MPPI controller is the negative mean of the cost $c_{k,t}$ when executing the controller $\mathbf{m}_k$. Then the RL policy parameters $\theta$ are updated as
$
	\theta \leftarrow \theta + \alpha \nabla_\theta J(\theta)
$
with a learning rate $\alpha>0$.


\section{Approach: a Meta RL-Based Adaptable and Generalizable Optimizer}
\begin{figure}[t]
\centering
\includegraphics[width=1.0\columnwidth]{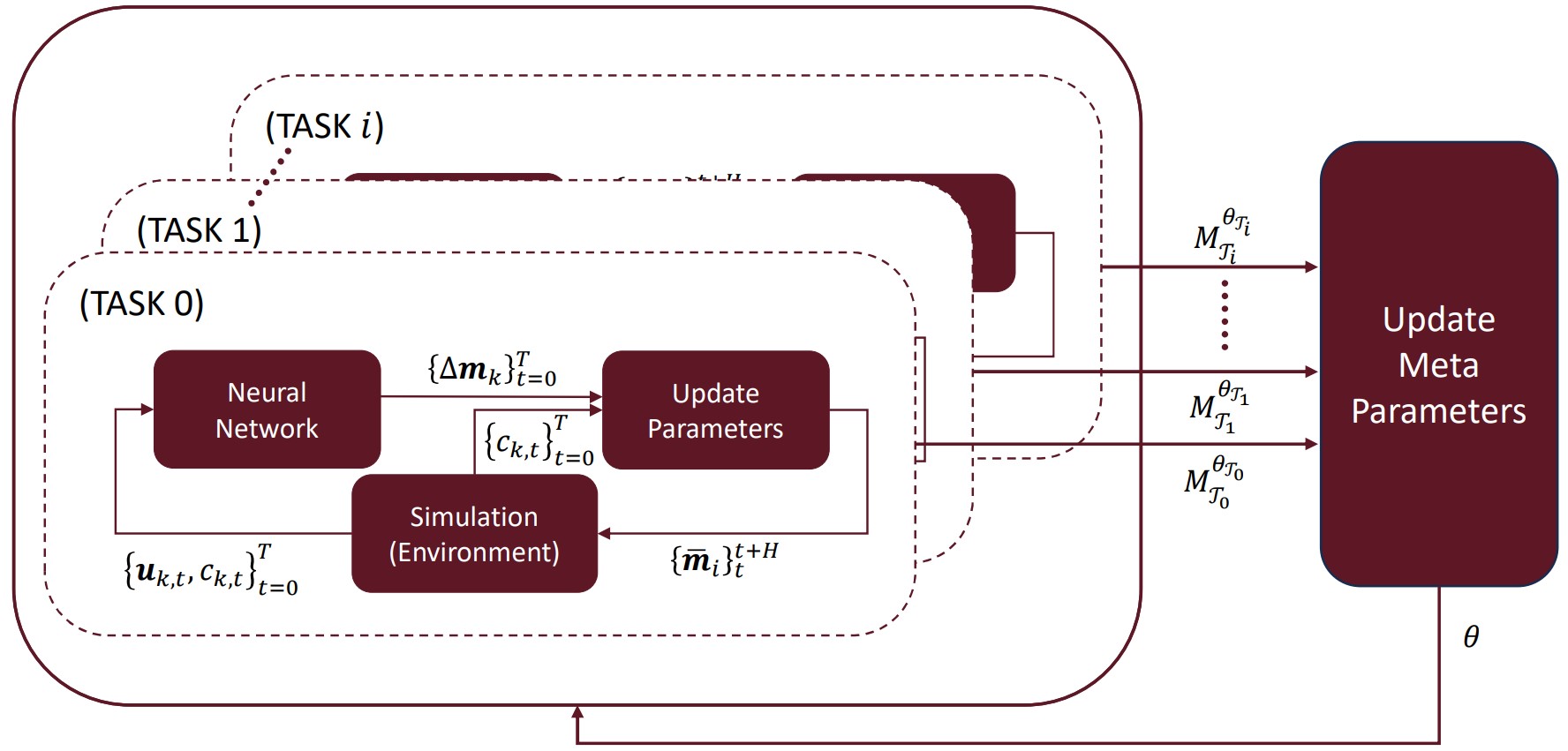}
\caption{Meta-RL Based Learnable Optimizer.}
\label{fig:conops}
\end{figure}

To enable optimizer with adaptation and generalizability in various tasks, we generalize the above approach under a meta-RL lens; illustrated in Figure \ref{fig:conops}. When adapting to a task $\mathcal{T}_i$ under the distribution $p(\mathcal{T})$, the meta optimizer $\theta$ becomes $\theta_{\mathcal{T}_i}$ with one or multiple gradient ascent updates with $\beta>0$ as 
\begin{equation}\label{eq:meta}
    \theta_{\mathcal{T}_i} = \theta + \beta \nabla_\theta J_{\mathcal{T}_i}(\theta),
\end{equation}
with the task-specific return $J_{\mathcal{T}_i}$. To update the meta parameter $\theta$, we build the loss function across multiple tasks followed by gradient-based update (with $\gamma>0$) as 
\begin{equation}\label{eq:meta-l2o}
\begin{aligned}
    J_{p(\mathcal{T})}(\theta) &= \sum_{\mathcal{T}_i\sim p(\mathcal{T})} J_{\mathcal{T}_i}(\theta_{\mathcal{T}_i} ) \\
    \theta &\leftarrow \theta + \gamma \nabla_\theta J_{p(\mathcal{T})}(\theta).
\end{aligned}
\end{equation}
The full algorithm of the generic meta-optimizer is described in Algorithm \ref{alg:gmo}.




\begin{algorithm}[tb]
\caption{Generic Meta-Optimizer for MPPI}
\label{alg:gmo}
\textbf{Require}: Initial controller $\mathbf{m}_{0}$, meta optimizer $\theta$, task distribution $p(\mathcal{T})$, learning rates $\beta, \gamma$ \\
\textbf{Output}: Optimal meta-optimizer  $\theta^{\ast}$

\begin{algorithmic}[1] 
\WHILE{not done}
\STATE Sample task batch $\mathcal{T}_i\sim p(\mathcal{T})$
\FOR{each $\mathcal{T}_i$}
\STATE Sample $K$ trajectories $M_{\mathcal{T}_i}^{\theta} = \{\mathbf{m}_{i,h}\}_{h=0}^{H}$ with $\theta$.
\FOR{each controller $\mathbf{m}_{i,h}$}
    \STATE Execute $\mathbf{m}_{i,h}$ and collect $\mathcal{D}_{i,k} = \{\mathbf{u}_{k,t},c_{k,t}\}_{t=0}^T$    
\ENDFOR

\STATE Calculate $\nabla_\theta J_{\mathcal{T}_i}(\theta)$ using  Eq. \eqref{eq:pm_grad_fnc}
\STATE Update $\theta_{\mathcal{T}_i}$ using Eq. \eqref{eq:meta}
\STATE Sample trajectories $M_{\mathcal{T}_i}^{{\theta}_{\mathcal{T}_i}} = \{\mathbf{m}_{i,h}\}_{h=0}^{H}$ with $\theta_{\mathcal{T}_i}$

\ENDFOR
\STATE Update meta optimizer using Eq. \eqref{eq:meta-l2o} using each $M_{\mathcal{T}_i}^{{\theta}_{\mathcal{T}_i}}$
\ENDWHILE
\end{algorithmic}
\end{algorithm}

\section{Experiments}
\begin{figure}[t]
\centering
\includegraphics[width=1.0\columnwidth]{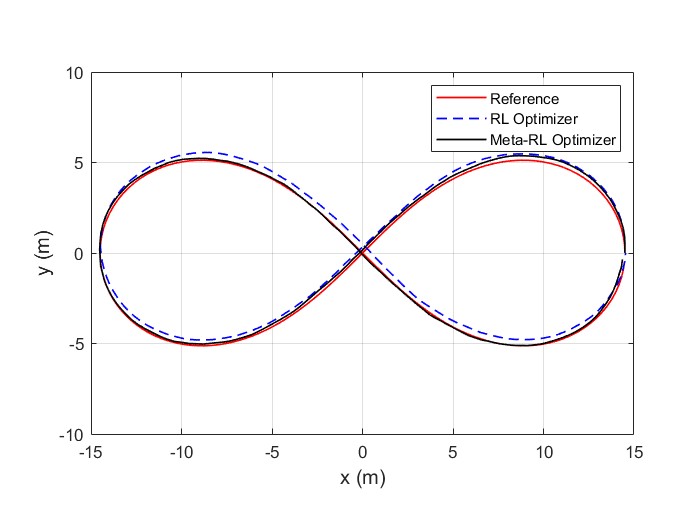}
\caption{Trajectory Comparison.}
\label{fig:trj_comp}
\end{figure}

\begin{figure}[t]
\centering
\includegraphics[width=1.0\columnwidth]{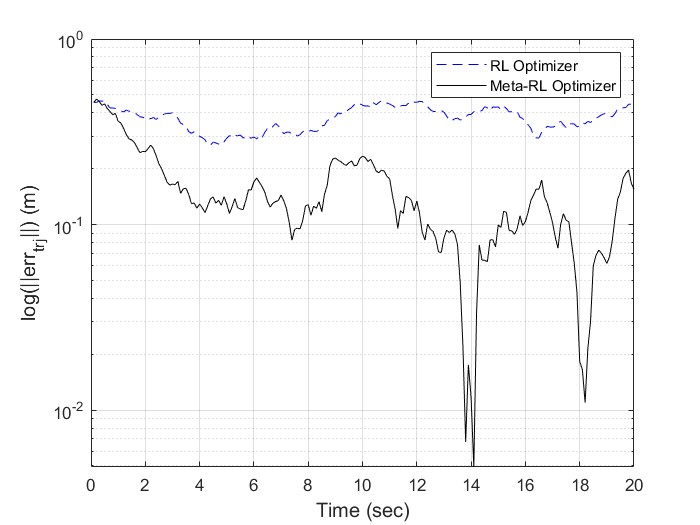}
\caption{Trajectory Error.}
\label{fig:trj_err}
\end{figure}

The experiment for the performance verification of the meta-RL optimizer is to compare the results of the path-following of MPPI including in the RL optimizer and the proposed optimizer, respectively. As shown in Figures \ref{fig:trj_comp} and \ref{fig:trj_err}, the performance of MPPI using the meta-RL optimizer is better than that using the RL optimizer.

\section{Conclusion and Future Works}
We propose a meta-RL-based approach to learn optimizers that are \textit{adaptable} and \textit{generalizable} to solving new control tasks. Future works will focus on solving constrained and hybrid (both discrete/continuous variables) problems.

\bibliography{aaai24}


\end{document}